\documentclass[10pt,twocolumn]{article}
\usepackage[numbers,sort&compress]{natbib}
\usepackage[hidelinks]{hyperref}
\usepackage[margin=0.75in]{geometry} 
\usepackage{times}
\usepackage{microtype}
\usepackage{balance} 

\usepackage{graphicx}
\usepackage{caption}
\usepackage{subcaption} 
\usepackage{booktabs}   
\usepackage{colortbl}
\usepackage{xcolor}

\definecolor{best}{rgb}{0.96, 0.62, 0.62}    
\definecolor{second}{rgb}{0.98, 0.78, 0.58}  
\definecolor{third}{rgb}{0.99, 0.92, 0.70}   
\usepackage{amsmath,amssymb}

\usepackage[hidelinks]{hyperref}

\title{GVGS: Gaussian Visibility-Aware Multi-View Geometry for Accurate Surface Reconstruction}
\author{
Mai Su$^{1}$,
Qihan Yu$^{1}$,
Zhongtao Wang$^{1}$,
Yilong Li$^{1}$,
Chengwei Pan$^{2}$,
Yisong Chen$^{1}$,
Guoping Wang$^{1,*}$
Fei Zhu$^{1,*}$\\
$^{1}$School of Computer Science, Peking University\\
$^{2}$Institute of Artificial Intelligence, Beihang University
}

\date{} 

\newcommand{\FirstPageTeaser}{%
  \vspace{-0.6em}
  \begin{center}
    \begin{minipage}{0.98\textwidth}
      \centering
      \includegraphics[width=\textwidth]{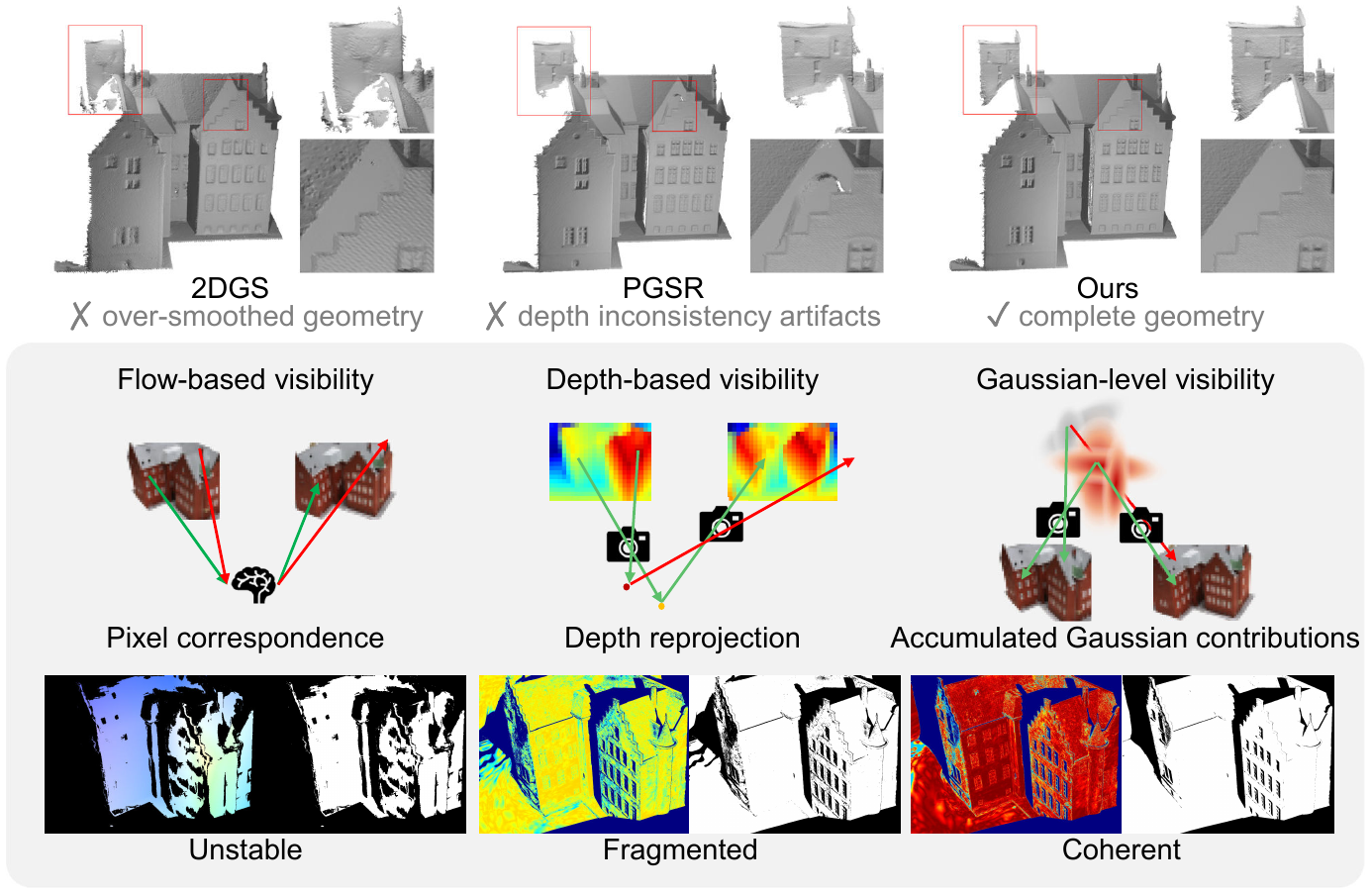}

      \captionsetup{type=figure}
\caption{Modeling visibility is fundamental for accurate surface reconstruction.
We compare three visibility estimation paradigms: flow-based correspondence, depth-based reprojection, and our Gaussian-level aggregation.
Flow-based methods rely on pixel correspondences and are prone to noisy and unstable matches, while depth-based methods infer visibility via reprojection, often leading to fragmented and incomplete supervision due to depth inaccuracies.
In contrast, our Gaussian-level formulation aggregates cross-view Gaussian contributions to produce a coherent visibility signal over co-visible regions.
As illustrated in the top row, unreliable visibility leads to over-smoothed geometry (2DGS~\cite{huang20242d}) and depth artifacts (PGSR~\cite{chen2024pgsr}), whereas our method enables complete and consistent reconstruction.
Flow visualizations depict 2D pixel motions (correspondences), while heatmaps (Depth and Ours) use warmer colors to indicate higher visibility intensity.
}
  \label{fig:teaser}
    \end{minipage}
  \end{center}
  \vspace{0.6em}
}

\begin{document}

\twocolumn[
  \maketitle
  \FirstPageTeaser
]

\begin{abstract}
3D Gaussian Splatting (3DGS) has emerged as a highly efficient representation for novel view synthesis. However, extracting accurate surfaces from 3DGS remains challenging due to unreliable geometric supervision. Existing methods heavily rely on depth-based reprojection for visibility estimation and multi-view consistency. This creates a fundamental circular dependency: precise visibility estimation requires accurate depth, yet depth supervision itself is conditioned on visibility. To break this cycle, we rethink multi-view geometric supervision through the lens of visibility, modeling it directly at the Gaussian level. Specifically, we propose a Gaussian Visibility-aware Multi-View (GVMV) geometric consistency formulation. By aggregating the cross-view visibility of shared Gaussians, GVMV enables robust supervision specifically over co-visible regions. Furthermore, to effectively integrate monocular depth priors, we introduce a progressive Quadtree-calibrated Depth Constraint (QDC). Guided by our visibility formulation, QDC performs block-wise affine calibration to mitigate scale ambiguity while strictly preserving local geometric structures. Extensive experiments on the DTU and Tanks and Temples benchmarks demonstrate that our approach consistently outperforms prior Gaussian-based methods in surface reconstruction accuracy.
Our code is fully open-sourced and available at an anonymous repository \url{https://github.com/GVGScode/GVGS}.
\end{abstract}

\section{Introduction}
3D Gaussian Splatting (3DGS)~\citep{kerbl20233d} has revolutionized novel view synthesis by delivering high-fidelity rendering at real-time speeds, significantly outperforming implicit NeRF representations~\citep{mildenhall2021nerf,barron2021mip,muller2022instant} in computational efficiency. By directly optimizing geometric and appearance parameters via rasterization, 3DGS serves as a strong foundation for high-fidelity rendering. However, because the representation is inherently optimized for rendering rather than geometric fidelity, accurately extracting surface geometry from 3DGS remains a challenging problem.

Under purely photometric supervision, the unstructured, volumetric nature of Gaussian primitives may cause them to drift from true surfaces while still explaining image observations. This leads to geometric ambiguity, thickness artifacts, and degraded multi-view consistency~\citep{xiao2025mcgs}. To mitigate this, recent efforts incorporate geometry-aware regularizations~\citep{turkulainen2025dn,chen2024vcr}, reformulate Gaussians as surface-aligned 2D elements~\citep{chen2024pgsr,guedon2024sugar,huang20242d}, or combine them with implicit fields~\citep{yu2024gsdf,jiang2025geometry}. These advances underscore the necessity of embedding geometric structure into Gaussian representations.

Despite these efforts, existing methods primarily rely on depth-based supervision, operating under the assumption that visibility can be reliably inferred from depth reprojection. When depth estimates become unreliable—such as under occlusions, wide baselines, or weak textures—both visibility and geometric constraints degrade simultaneously. While some approaches leverage monocular depth or normal priors to guide optimization~\citep{liang2024gs,turkulainen2025dn}, these priors suffer from scale ambiguity and local inconsistency. Without addressing the underlying visibility bottleneck, simply enforcing multi-view depth consistency alongside monocular priors often oversmooths fine structures or introduces new artifacts.

In this paper, we break this circular dependency by revisiting multi-view geometric supervision through the lens of Gaussian-level visibility. We propose a Gaussian Visibility-aware Multi-View (GVMV) consistency formulation. Instead of relying on depth reprojection, our method estimates per-Gaussian visibility across views by aggregating rendering contributions, thereby constructing a robust, visibility-aware supervision signal. This mechanism enables reliable geometric consistency to be enforced over a broader set of co-visible regions, bypassing the limitations of traditional depth reprojection.

To effectively integrate monocular priors without compromising fine-grained structures, we further introduce a progressive Quadtree-calibrated Depth Constraint (QDC). By performing coarse-to-fine, block-wise affine calibration under visibility-aware guidance, QDC mitigates scale ambiguity while preserving local geometric fidelity, enabling monocular depth to serve as a highly effective geometric prior.

In summary, our contributions are threefold:
\begin{itemize}

\item \textbf{A new paradigm for multi-view geometric supervision.}
We shift from pixel-aligned depth consistency to Gaussian-centric visibility reasoning, 
reformulating supervision from image space to primitive space for more robust and 
physically grounded geometry. 
This perspective resolves the circular dependency in depth-based methods by decoupling visibility from depth reprojection.

\item \textbf{Gaussian visibility-aware multi-view geometric formulation.}
We propose GVMV, a Gaussian-level framework that explicitly captures cross-view co-visibility, 
enabling robust geometric consistency beyond depth-reliable regions.

\item \textbf{Visibility-guided monocular depth alignment strategy.}
We introduce QDC, a progressive quadtree-calibrated alignment strategy that integrates monocular priors 
to improve both global structural consistency and local geometric fidelity.

\end{itemize}

\section{Related Works}
\paragraph{Novel View Synthesis}
Neural radiance field (NeRF)-based methods model scenes as continuous volumetric fields optimized via differentiable rendering and have been widely used for novel view synthesis~\cite{mildenhall2021nerf, gu2020cascade}, but incur high computational cost due to dense ray marching and neural network evaluations~\cite{barron2021mip}.
As an efficient alternative, 3D Gaussian Splatting (3DGS) represents scenes using anisotropic Gaussian primitives and performs rendering via rasterization~\cite{kerbl20233d}, enabling fast optimization and real-time performance.
Building on this representation, subsequent works have advanced rendering quality and efficiency~\cite{yu2024mip, fang2024mini}, and broadened its applicability to large-scale~\cite{kerbl2024hierarchical, su2025hug}, dynamic~\cite{luiten2024dynamic, duan20244d}, and sparse-view scenarios~\cite{chen2024mvsplat, zhang2025transplat}.

\paragraph{Gaussian Splatting for Surface Reconstruction}
To explicitly recover surface geometry from Gaussian splatting, prior work reformulates Gaussian primitives into surface-oriented representations.
SuGaR~\cite{guedon2024sugar} introduces a surface-alignment regularizer that encourages Gaussians to form locally surface-tangent configurations, and derives an approximate distance function for efficient mesh extraction via Poisson reconstruction~\cite{kazhdan2006poisson}.
Planar-based formulations reinterpret anisotropic Gaussians as locally planar primitives, enabling unbiased depth and normal rendering and facilitating geometry-aware constraints~\cite{chen2024pgsr}.
More generally, subsequent works constrain Gaussians to planar or disk-like configurations to better adhere to underlying surfaces~\cite{huang20242d,yang2025introducing,zhang2025quadratic}.
In parallel, recent approaches address transparent surface reconstruction by learning transparency attributes to handle view-dependent appearance and geometric ambiguities~\citep{li2025tsgs}.

Another line of work combines Gaussian splatting with implicit geometry fields to leverage the complementary strengths of explicit and implicit representations.
Methods such as GSDF and GaussianUDF jointly optimize Gaussian primitives with signed or unsigned distance functions to guide surface reconstruction while retaining efficient rendering~\cite{yu2024gsdf,li2025gaussianudf}.
Geometry Field Splatting further unifies this paradigm by representing geometry fields with Gaussian surfels and deriving an efficient differentiable rendering formulation, establishing a more principled connection between Gaussian splatting and implicit geometry~\cite{jiang2025geometry}.
GOF models surfaces via a compact continuous opacity field over Gaussian primitives, enabling efficient and memory-compact reconstruction in unbounded scenes without explicit distance fields or dense volumetric sampling~\cite{yu2024gaussian}.

\paragraph{Geometric Constraints for Gaussian-based Surface Reconstruction}
Beyond representation-level designs, recent work improves Gaussian-based surface reconstruction by incorporating geometric constraints during optimization.
2DGS~\cite{huang20242d} introduces a normal consistency constraint to explicitly model per-Gaussian surface normals and enforce alignment with depth-derived normals, stabilizing disk-like primitives.
DN-Splatter~\cite{turkulainen2025dn} aligns Gaussian orientations with monocular normal priors and further imposes local smoothness to enforce consistency of depth, normal, and scale among neighboring Gaussians.
PGSR~\cite{chen2024pgsr} proposes a multi-view geometric consistency loss that jointly enforces depth and photometric consistency across views by constraining shared planar Gaussian structures, drawing inspiration from classical multi-view stereo formulations~\cite{campbell2008using}.
These methods highlight the importance of geometric regularization for accurate surface reconstruction, but typically assume reliable depth estimation or consistent cross-view scale.

Despite their effectiveness, existing multi-view constraints still rely heavily on accurate Gaussian depth and remain sensitive to depth bias in challenging regions.
Moreover, monocular supervision does not explicitly address cross-view scale ambiguity.
To address these limitations, we explicitly model cross-view visibility at the Gaussian level and introduce a quadtree-calibrated monocular depth constraint, enabling more robust multi-view supervision and reliable single-view depth guidance under imperfect depth priors.

\section{Method}
\begin{figure*}
    \centering
    \includegraphics[width=0.98\linewidth]{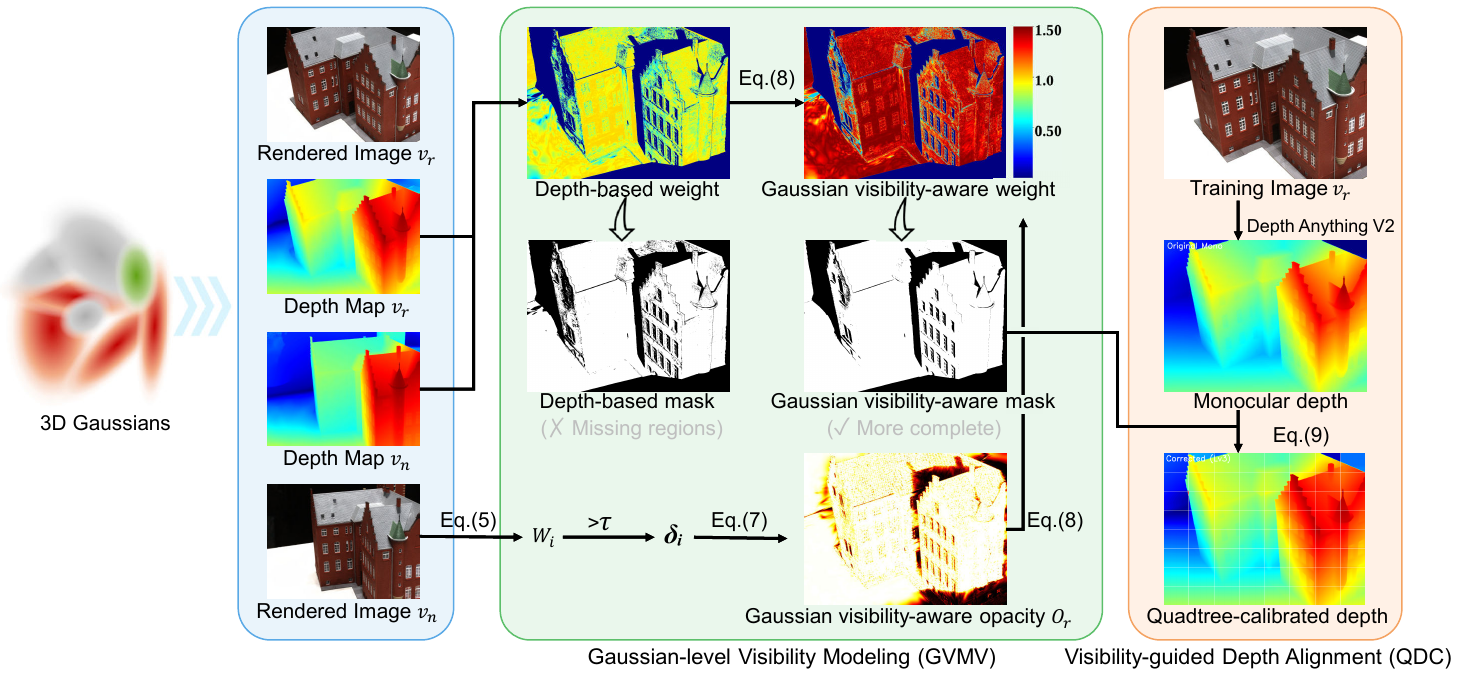}
\caption{
Overview of the GVGS framework.
Unlike conventional depth-based supervision that inherently yields incomplete visibility masks, we explicitly characterize cross-view visibility at the Gaussian level. 
Given a reference view $v_r$ and a neighboring view $v_n$, we compute per-Gaussian visibility weights $W_i$ in $v_n$ (Eq.~\eqref{eq:gaussian_visibility_weight}) to derive binary indicators $\delta_i$. 
Projecting these indicators back to $v_r$ constructs a visibility-aware opacity mask $O_r$ (Eq.~\eqref{eq:accOpacity}). 
This provides a robust and comprehensive mask over co-visible regions, driving our geometric consistency loss $L_{\text{gvmvgeom}}$ (Eq.~\eqref{eq:gvmvgeom}). 
Concurrently, our QDC progressively aligns monocular depth priors (Depth-Anything V2) with Gaussian-rendered depth under visibility guidance to formulate $L_{\text{qdc}}$ (Eq.~\eqref{eq:qta_loss}). Heatmaps for \textit{depth-based} and \textit{visibility-aware} weights share the same color bar.}

\label{fig:overview}
\end{figure*}

Given calibrated multi-view RGB images, our framework reconstructs scene geometry via 3D Gaussian Splatting by synergizing multi-view geometric cues with monocular depth priors (Fig.~\ref{fig:overview}). Our method introduces two core components:
(1) Gaussian Visibility-aware Multi-View consistency (GVMV), which explicitly models cross-view visibility at the Gaussian level to provide robust geometric supervision; and
(2) a Quadtree-calibrated Depth Constraint (QDC), which refines monocular depth to offer coherent structural guidance.

In the following, we briefly review the 3DGS preliminary before detailing our proposed GVMV, QDC, and the joint optimization objective.

\subsection{Preliminary: 3D Gaussian Splatting}
\label{sec:Preli}
3D Gaussian Splatting (3DGS)~\cite{kerbl20233d} represents scenes using a set of learnable 3D Gaussian primitives. 
Each Gaussian is characterized by a center $\boldsymbol{\mu} \in \mathbb{R}^3$, an anisotropic covariance matrix $\boldsymbol{\Sigma} \in \mathbb{R}^{3 \times 3}$, an opacity $\alpha \in [0,1]$, and appearance features for view-dependent color. 
The spatial influence of a Gaussian at a 3D location $\boldsymbol{x}$ is defined as:
\begin{equation}
G(\boldsymbol{x}) = \exp\left( -\tfrac{1}{2} (\boldsymbol{x} - \boldsymbol{\mu})^\top \boldsymbol{\Sigma}^{-1} (\boldsymbol{x} - \boldsymbol{\mu}) \right).
\label{eq1:gsvalue}
\end{equation}

During rasterization, these 3D primitives are projected onto the image plane via the camera model, yielding 2D Gaussian footprints with projected center $\boldsymbol{\mu}'$ and covariance $\boldsymbol{\Sigma}'$:
\begin{equation}
G'(\boldsymbol{x}') = \exp\left( -\tfrac{1}{2} (\boldsymbol{x}' - \boldsymbol{\mu}')^\top \boldsymbol{\Sigma}'^{-1} (\boldsymbol{x}' - \boldsymbol{\mu}') \right),
\label{eq2:gs2dproject}
\end{equation}
where $\boldsymbol{x}' \in \mathbb{R}^2$ denotes the pixel location. 
The final color $\boldsymbol{C}(\boldsymbol{x}')$ is computed by alpha-blending $N$ depth-sorted Gaussians overlapping the pixel:
\begin{equation}
\boldsymbol{C}(\boldsymbol{x}') = \sum_{i} T_i\, \alpha_i\, G'_i(\boldsymbol{x}')\, \boldsymbol{c}_i, \quad
T_i = \prod_{j=1}^{i-1} \left( 1 - \alpha_j\, G'_j(\boldsymbol{x}') \right),
\label{eq3:gscolor}
\end{equation}
where $\boldsymbol{c}_i$ is the color of the $i$-th Gaussian, $T_i$ denotes the accumulated transmittance along the ray up to the $i$-th Gaussian.

\subsection{Gaussian Visibility-Aware Multi-View Geometric Consistency}
\label{sec:GVMVGEOM}

To break aforementioned circular dependency, we establish a core principle: 
\textbf{multi-view consistency should be enforced over all co-visible regions}, 
rather than being limited to depth-reliable pixels.

To formalize this concept, we define co-visibility at the Gaussian level. 
Given two views $v_r$ and $v_n$, a Gaussian $g_i$ is considered co-visible 
if it has non-zero visibility probability in both views:
\begin{equation}
\mathcal{C} = \{ g_i \mid p_i^{(v_r)} \cdot p_i^{(v_n)} > 0 \}.
\end{equation}
Unlike prior methods that infer visibility from depth or patch consistency~\cite{5226635,1640800}, 
\textbf{we model visibility at the Gaussian level}. 
Existing methods operate on pixel-aligned depth consistency and implicitly rely on accurate depth, 
whereas our formulation derives visibility directly from volumetric compositing, decoupling it from depth quality. 
This enables more robust supervision, especially in regions where depth reprojection is unreliable.

Guided by this insight, we introduce a \textbf{Gaussian-based visibility estimation} module to explicitly capture cross-view co-visibility. 
This serves as the foundation of our \textbf{Gaussian visibility-aware multi-view geometric consistency} formulation (overviewed in Fig.~\ref{fig:overview}), 
which dynamically enforces geometric supervision over all identified co-visible regions.

\paragraph{Gaussian-based Visibility Estimation.}
Given a reference view $v_r$ and a neighboring view $v_n$, our objective is to identify which Gaussian primitives are visible in $v_n$ and should therefore contribute to cross-view supervision.

To this end, we explicitly calculate the rendering contribution of each Gaussian during the differentiable rasterization of the neighboring view $v_n$. 
We define the visibility weight $W_i \in \mathbb{R}^+$ of each Gaussian based on its cumulative contribution to the rendered image:
\begin{equation}
W_i
=
\sum_{\mathbf{x} \in \Omega_n}
\alpha_i(\mathbf{x}) \cdot T_i(\mathbf{x}),
\label{eq:gaussian_visibility_weight}
\end{equation}
where $\Omega_n$ denotes the image domain of view $v_n$, and $T_i(\mathbf{x})$ represents the accumulated transmittance.

We reinterpret $W_i$ as the \emph{expected contribution} of Gaussian $g_i$ to the rendered image under volumetric compositing. 
From a probabilistic perspective, $W_i$ can be viewed as a Monte Carlo estimate of the likelihood that $g_i$ is observed along camera rays, as it aggregates its contributions across all pixels in the image domain.

Based on this interpretation, we define the visibility of each Gaussian as a Bernoulli random variable:
\begin{equation}
\delta_i \sim \mathrm{Bernoulli}(p_i), \quad
p_i = \frac{W_i}{\sum_j W_j},
\end{equation}
where $p_i$ represents the normalized visibility probability of Gaussian $g_i$ in view $v_n$. In practice, for computational efficiency and robustness, we adopt a thresholded approximation of this probabilistic formulation:
$\delta_i = \mathbb{I}(W_i > \tau)$,
where $\tau$ is a small threshold used to suppress negligible contributions. 
This binary approximation provides a stable and efficient estimate of visibility while remaining consistent with the underlying probabilistic interpretation.




\paragraph{Visibility Projection to the Reference View.}
We subsequently transfer this estimated visibility back to the reference view $v_r$ to construct a selectively accumulated opacity map:
\begin{equation}
O_r(\mathbf{x})
=
\sum_{i} \delta_i \, \alpha_i(\mathbf{x})
\prod_{j<i} \left(1 - \alpha_j(\mathbf{x}) \right),
\label{eq:accOpacity}
\end{equation}
where $\delta_i$ serves as a visibility gate, activating solely the Gaussians dynamically confirmed as visible in $v_n$.

Consequently, $O_r(\mathbf{x})$ aggregates contributions strictly from Gaussians that are co-visible across the two views while preserving the standard depth-ordered alpha compositing. Importantly, $O_r(\mathbf{x})$ transcends simple opacity accumulation; it acts as a visibility-aware weighting term at the Gaussian level that essentially encodes cross-view co-visibility. This design remains reliable even when depth-based reprojection fails.

\paragraph{Gaussian Visibility-Aware Geometric Consistency.}
We build upon the multi-view geometric consistency loss $L_{\text{mvgeom}}$ introduced in PGSR~\cite{chen2024pgsr}. In their formulation, each reference pixel $\mathbf{x}$ is associated with a forward--backward reprojection error $\phi(\mathbf{x})$, which is converted into a confidence weight via a monotonic function ($\exp(-\phi(\mathbf{x}))$). Pixels exhibiting large reprojection errors ($\phi(\mathbf{x}) > 1$) are typically excluded from supervision.

We fundamentally extend this paradigm by incorporating our Gaussian-level visibility prior through $O_r(\mathbf{x})$. This unlocks geometric supervision across a substantially broader set of co-visible regions, rather than being restricted to areas endowed with initially reliable depth reprojection. As highlighted in Fig.~\ref{fig:overview}, regions that were previously unsupervised (e.g., textureless facade areas) receive robust geometric constraints through our GVMV formulation.

The resulting Gaussian visibility-aware multi-view geometric consistency loss is thus defined as:
\begin{equation}
L_{\mathrm{gvmvgeom}}
=
\frac{1}{|\mathcal{V}|}
\sum_{\mathbf{x} \in \mathcal{V}}
\left(
\exp\!\bigl(-\phi(\mathbf{x})\bigr)
+ \lambda \, O_r(\mathbf{x})
\right)
\, \phi(\mathbf{x}),
\label{eq:gvmvgeom}
\end{equation}
where $\mathcal{V}$ denotes the union of two subsets: (i) pixels satisfying conventional depth-based consistency, and (ii) pixels identified as co-visible by the proposed Gaussian visibility-aware opacity map $O_r$. The hyperparameter $\lambda$ dynamically balances the contribution of the visibility term.

\subsection{Quadtree-calibrated Monocular Depth Constraint}
\label{sec:QCD}

While recent large vision models provide highly detailed monocular depth priors~\cite{yang2024depth,xu2025pixel}, their integration into multi-view pipelines is bottlenecked by \emph{scale ambiguity} and \emph{view-dependent bias}. Existing methods typically mitigate this via global scale-and-shift calibration using sparse SfM points or by restricting supervision strictly to multi-view consistent pixels~\cite{kerbl2024hierarchical,gao2025citygs}. However, such global strategies inherently fail to capture spatially varying depth distortions.

To resolve this, we introduce a \textbf{progressive quadtree-calibrated depth constraint} strategy. Bypassing rigid global transformations, we hierarchically align monocular depth with Gaussian-rendered depth across multiple spatial scales. Crucially, this progressive local calibration is dynamically guided by reliable co-visible regions identified by our visibility-aware formulation.

\paragraph{Coarse-to-fine Quadtree Alignment.}
During training, we employ a coarse-to-fine quadtree schedule.At iteration $t$, we set the quadtree level as $L(t)=\left\lfloor \tfrac{t}{\Delta t} \right\rfloor$, and partition the image space into $2^{L(t)} \times 2^{L(t)}$ uniform blocks. As training progresses, $L(t)$ gradually increases, smoothly transitioning the depth alignment from a global coarse calibration to a fine-grained local refinement.

For each quadtree block $\mathcal{B}_k$, we align the monocular depth $D_m(\mathbf{x})$ with the Gaussian-rendered depth $D_g(\mathbf{x})$ using a block-wise affine model~\cite{ranftl2020towards}:
\begin{equation}
\begin{aligned}
D_m'(\mathbf{x})
&=
a_k \, D_m(\mathbf{x}) + b_k,
\qquad
\mathbf{x} \in \mathcal{B}_k \cap \mathcal{V}
, \\
a_k
&=
\frac{\sigma_{\mathbf{y}\in\mathcal{B}_k}\!\bigl(D_g(\mathbf{y})\bigr)}
     {\sigma_{\mathbf{y}\in\mathcal{B}_k}\!\bigl(D_m(\mathbf{y})\bigr)}, \\
b_k
&=
\mu_{\mathbf{y}\in\mathcal{B}_k}\!\Bigl(
D_g(\mathbf{y}) - a_k\, D_m(\mathbf{y})
\Bigr),
\end{aligned}
\label{eq:qta_affine}
\end{equation}
where $(a_k, b_k)$ denote the affine scale and shift parameters associated with block $\mathcal{B}_k$. To ensure robustness against outliers, $\mu(\cdot)$ and $\sigma(\cdot)$ denote robust estimators for location and scale, implemented as the median and the median absolute deviation (MAD), respectively.

Intuitively, this hierarchical formulation adapts to locally varying depth distortions. Coarse quadtree levels correct fundamental global scale mismatches, while finer levels dynamically capture and rectify spatially varying biases. Specifically, this calibration is exclusively restricted to pixels within $\mathcal{V}$. This guarantees that the alignment is anchored solely by reliable geometric cues.

\begin{figure}[!t]
    \centering
    \includegraphics[width=\linewidth]{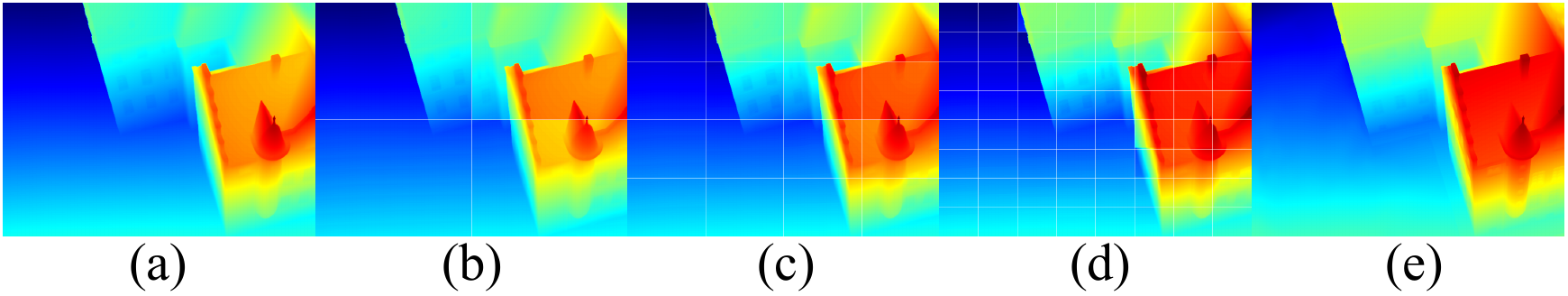}
    \caption{Progressive quadtree depth calibration.
(a) Raw monocular depth and (e) Gaussian-rendered depth exhibit clear misalignment.
(b--d) Coarse-to-fine block-wise affine calibration (Lv1--Lv3) progressively aligns monocular depth with Gaussian-rendered depth.}
    \label{fig:quadtree_depth_alignment}
\end{figure}

To maximize efficiency, the affine parameters $(a_k, b_k)$ are only recomputed when the quadtree level $L(t)$ increments; they are cached and reused across intermediate iterations, resulting in negligible computational overhead. As illustrated in Fig.~\ref{fig:quadtree_depth_alignment}, this progressive strategy yields increasingly precise alignment, effectively recovering regions burdened with large initial depth biases.

\paragraph{Quadtree-calibrated Monocular Depth Constraint.}
Following alignment, we penalize the discrepancy between the calibrated monocular depth $D_m'(\mathbf{x})$ and the Gaussian-rendered depth $D_g(\mathbf{x})$ using an $\ell_1$ loss to enforce geometric supervision:
\begin{equation}
L_{{qdc}}
=
\sum_{\mathbf{x} \in \mathcal{V}}
\left\|
D_m'(\mathbf{x})
-
D_g(\mathbf{x})
\right\|_1 ,
\label{eq:qta_loss}
\end{equation}
where $\mathcal{V}$ denotes the trusted region defined in Eq.~\eqref{eq:gvmvgeom}.

We apply this monocular depth constraint primarily during the early stages of training to provide coarse structural guidance, rather than serving as a rigid final supervision signal. By restricting supervision exclusively to visibility-aware regions and utilizing coarse-to-fine local calibration, our formulation transforms raw monocular depth into a stable, highly accurate geometric prior. This synergy enhances both global structural consistency and local geometric fidelity, remaining inherently robust to noisy or biased depth predictions.

\subsection{Training Objective}
\label{sec:trainingobj}
We optimize our proposed framework end-to-end using a joint loss function, formulated as a weighted sum of multiple objective terms:
\begin{equation}
L
=
L_{\mathrm{rgb}}
+
L_s
+
L_{\mathrm{mvrgb}}
+
L_{\mathrm{gvmvgeom}}
+
L_{\mathrm{qdc}} .
\end{equation}

Specifically, $L_{\text{rgb}}$ is the standard 3DGS photometric loss (combining $\ell_1$ and SSIM). Following PGSR~\cite{chen2024pgsr}, $L_s$ imposes single-view depth and normal regularizations, and $L_{\text{mvrgb}}$ enforces multi-view photometric consistency. Building upon these baselines, our proposed terms, $L_{\text{gvmvgeom}}$ and $L_{\text{qdc}}$, explicitly establish robust visibility-aware multi-view geometric supervision and dynamically calibrate monocular depth priors, respectively.

\begin{figure*}[!t]
    \centering
    \includegraphics[width=\textwidth,height=0.70\textheight,keepaspectratio]{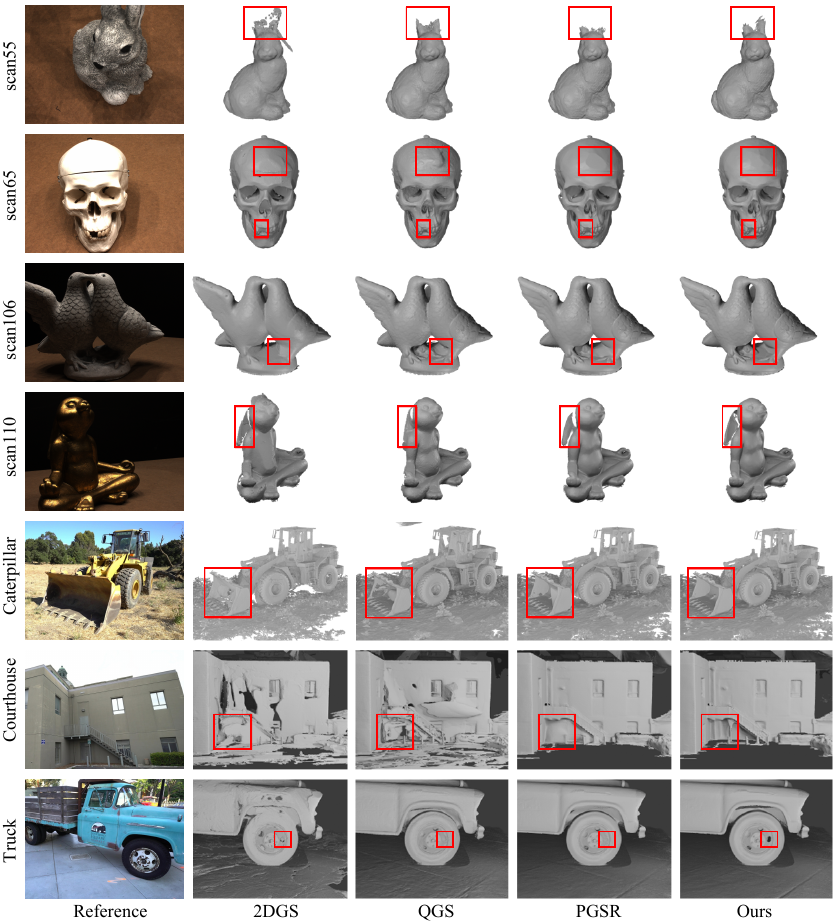}
    \caption{Qualitative comparison of reconstructed geometry with related methods on DTU and Tanks and Temples.
Red boxes highlight regions with noticeable geometric differences.}
    \label{fig:geocompare}
\end{figure*}

\begin{table*}[!ht]
\centering
\small
\setlength{\tabcolsep}{4pt}
\renewcommand{\arraystretch}{1.05}
\caption{\textbf{Chamfer Distance on DTU (lower is better).}
For each scan, the best, second, and third results are highlighted with \textcolor{best}{red}, \textcolor{second}{orange}, and \textcolor{third}{yellow} backgrounds, respectively.
}
\label{tab:dtu_cd}
\resizebox{\textwidth}{!}{%
\begin{tabular}{l|ccccccccccccccc|cc}
\hline
CD (mm)$\downarrow$
& 24 & 37 & 40 & 55 & 63 & 65 & 69 & 83 & 97 & 105 & 106 & 110 & 114 & 118 & 122
& Mean & Time \\
\hline
NeuS~\cite{wang2021neus}
& 1.00 & 1.37 & 0.93 & 0.43 & 1.10 & 0.65 & 0.57 & 1.48 & 1.09 & 0.83 & 0.52 & 1.20 & 0.35 & 0.49 & 0.54 & 0.84 & $>$12h \\
VolSDF~\cite{yariv2021volume}
& 1.14 & 1.26 & 0.81 & 0.49 & 1.25 & 0.70 & 0.72 & 1.29 & 1.18 & 0.70 & 0.66 & 1.08 & 0.42 & 0.61 & 0.55 & 0.86 & $>$12h \\
Neuralangelo~\cite{li2023neuralangelo}
& \cellcolor{second}0.37 & 0.72 & \cellcolor{second}0.35 & \cellcolor{second}0.35 & 0.87 & \cellcolor{second}0.54 & 0.53 & 1.29 & \cellcolor{third}0.97 & 0.73 & \cellcolor{second}0.47 & 0.74 & \cellcolor{second}0.32 & \cellcolor{third}0.41 & 0.43 & 0.61 & $>$128h \\
\hline
3DGS~\cite{kerbl20233d}
& 2.14 & 1.53 & 2.08 & 1.68 & 3.49 & 2.21 & 1.43 & 2.07 & 2.22 & 1.75 & 1.79 & 2.55 & 1.53 & 1.52 & 1.50 & 1.96 & 11.2min \\
SuGaR~\cite{guedon2024sugar}
& 1.47 & 1.33 & 1.13 & 0.61 & 2.25 & 1.71 & 1.15 & 1.63 & 1.62 & 1.07 & 0.79 & 2.45 & 0.98 & 0.88 & 0.79 & 1.33 & 1h \\
2DGS~\cite{huang20242d}
& 0.48 & 0.91 & 0.39 & 0.39 & 1.01 & 0.83 & 0.81 & 1.36 & 1.27 & 0.76 & 0.70 & 1.40 & 0.40 & 0.76 & 0.52 & 0.80 & 19.2min \\
GOF~\cite{yu2024gaussian}
& 0.50 & 0.82 & 0.37 & 0.37 & 1.12 & 0.74 & 0.73 & 1.18 & 1.29 & 0.68 & 0.77 & 0.90 & 0.42 & 0.66 & 0.49 & 0.74 & 1h \\
QGS~\cite{zhang2025quadratic}
& 0.42 & \cellcolor{third}0.65 & \cellcolor{third}0.36 & 0.37 & \cellcolor{third}0.85 & 0.65 & \cellcolor{third}0.50 & \cellcolor{third}1.14 & \cellcolor{third}0.97 & \cellcolor{third}0.61 & \cellcolor{third}0.48 & \cellcolor{third}0.67 & \cellcolor{third}0.34 & \cellcolor{third}0.41 & \cellcolor{third}0.37 & \cellcolor{third}0.59 & 48min \\
PGSR~\cite{chen2024pgsr}
& \cellcolor{third}0.39 & \cellcolor{second}0.54 & 0.39 & \cellcolor{third}0.36 & \cellcolor{best}0.78 & \cellcolor{third}0.57 & \cellcolor{second}0.49 & \cellcolor{second}1.07 & \cellcolor{second}0.64 & \cellcolor{second}0.59 & \cellcolor{second}0.47 & \cellcolor{second}0.54 & \cellcolor{best}0.30 & \cellcolor{second}0.37 & \cellcolor{second}0.34 & \cellcolor{second}0.52 & 40min \\
Ours
& \cellcolor{best}0.32 & \cellcolor{best}0.53 & \cellcolor{best}0.33 & \cellcolor{best}0.33 & \cellcolor{second}0.79 & \cellcolor{best}0.52 & \cellcolor{best}0.47 & \cellcolor{best}1.05 & \cellcolor{best}0.63 & \cellcolor{best}0.58 & \cellcolor{best}0.37 & \cellcolor{best}0.53 & \cellcolor{best}0.30 & \cellcolor{best}0.35 & \cellcolor{best}0.32 & \cellcolor{best}0.49 & 43min \\
\hline
\end{tabular}%
}
\end{table*}

\begin{table}[!t]
\caption{
\textbf{Quantitative F1-score comparison on the TNT dataset} (higher is better).
The best/second/third results are highlighted in \textcolor{best}{red}, \textcolor{second}{orange}, and \textcolor{third}{yellow}.
}
\label{tab:tnt_f1}
\centering
\small      
\setlength{\tabcolsep}{3pt}
\renewcommand{\arraystretch}{1.05}
\resizebox{\columnwidth}{!}{
\begin{tabular}{l|ccccccc}
\hline
F1-Score$\uparrow$
& N-angelo & 2DGS & GOF & PGSR & QGS & Ours(30k)& Ours(60k) \\
\hline
Barn
& \cellcolor{best}0.70
& 0.41
& 0.51
& \cellcolor{third}0.56
& 0.55
& 0.54
& \cellcolor{second}0.58 \\

Caterpillar
& 0.36
& 0.24
& \cellcolor{second}0.41
& \cellcolor{second}0.41
& \cellcolor{third}0.40
& 0.44
& \cellcolor{best}0.47 \\

Courthouse
& \cellcolor{best}0.28
& 0.16
& \cellcolor{best}0.28
& \cellcolor{second}0.26
& \cellcolor{best}0.28
& 0.22
& \cellcolor{third}0.24 \\

Ignatius
& \cellcolor{best}0.89
& 0.52
& 0.68
& \cellcolor{third}0.79
& \cellcolor{second}0.81
& 0.80
& \cellcolor{second}0.81 \\

Meetingroom
& \cellcolor{third}0.32
& 0.17
& 0.28
& \cellcolor{second}0.34
& 0.31
& 0.36
& \cellcolor{best}0.39 \\

Truck
& 0.48
& 0.45
& 0.58
& \cellcolor{second}0.65
& \cellcolor{third}0.64
& 0.68
& \cellcolor{best}0.68 \\

\hline
Mean
& \cellcolor{second}0.50
& 0.33
& \cellcolor{third}0.46
& \cellcolor{second}0.50
& \cellcolor{second}0.50
& 0.51
& \cellcolor{best}0.53 \\

Time
& $>$127h & 34min & 114min & 66min & 75min & 69min & 117min \\
\hline
\end{tabular}
}
\end{table}

\section{Experiments}
\subsection{Datasets}
We evaluate our method on two standard multi-view surface reconstruction benchmarks: \textbf{DTU} and \textbf{Tanks and Temples (TNT)}.
\textbf{DTU} comprises calibrated, object-centric scenes with high-fidelity structured-light ground truth. Following prior work~\cite{chen2024pgsr}, we report the symmetric Chamfer Distance on the standard 15 scans without explicit alignment to the ground-truth point clouds. We utilize the data preprocessed by 2DGS~\cite{huang20242d}.
\textbf{TNT} features diverse, large-scale indoor and outdoor environments. We evaluate on the six \emph{Intermediate} scenes (Barn, Caterpillar, Courthouse, Ignatius, Meetingroom, and Truck), reporting the per-scene F1-score. We adopt the data provided by GOF~\cite{yu2024gaussian} and strictly follow the evaluation protocol of QGS~\cite{zhang2025quadratic}.

\subsection{Implementation Details}

Our implementation builds upon PGSR~\cite{chen2024pgsr}, retaining its default hyperparameters for fair baseline comparison. We utilize monocular depth priors from Depth Anything V2~\cite{yang2024depth}, dynamically refined via our proposed QDC. Gaussian influence computation follows EAGLES~\cite{girish2024eagles}.

For DTU, scenes are optimized for 30k iterations. Our progressive quadtree split level $L(t)$ increments every $\Delta t=5k$ iterations. To provide robust structural guidance without over-constraining final convergence, $L_{\text{qdc}}$ is actively enforced between 7k and 25k iterations. For co-visibility estimation, negligible Gaussians are filtered using a threshold of $\tau = 1\text{e}{-4}$. We set the visibility weight $\lambda = 0.5$ (Eq.~\eqref{eq:gvmvgeom}). For TNT benchmark, training is extended to 60k iterations with hyperparameters scaled proportionally. Final meshes are extracted via TSDF fusion~\cite{newcombe2011kinectfusion}. All experiments are conducted using four NVIDIA RTX A6000 GPUs. Comprehensive configurations are provided in our released source code for reproducibility.

\subsection{Comparison}
We quantitatively compare our approach against both prominent implicit surface reconstruction methods~\cite{wang2021neus,yariv2021volume,li2023neuralangelo} and state-of-the-art Gaussian-based techniques~\cite{kerbl20233d,guedon2024sugar,huang20242d,yu2024gaussian,chen2024pgsr,zhang2025quadratic}. For fair comparison, we reproduce the results of PGSR~\cite{chen2024pgsr} and QGS~\cite{zhang2025quadratic} using their official implementations, while sourcing the remaining baseline results directly from QGS~\cite{zhang2025quadratic}.

For the \textbf{DTU} dataset, as detailed in Table~\ref{tab:dtu_cd}, our method achieves the lowest Chamfer Distance on \textbf{14 out of 15} evaluated scans. In terms of overall accuracy, it establishes a new state-of-the-art with a mean Chamfer Distance of \textbf{0.49 mm}, outperforming the best prior baseline by approximately \textbf{5\%}.

Qualitative comparisons in Fig.~\ref{fig:geocompare} further validate these quantitative gains. Our method recovers highly faithful geometric details, yielding more structurally complete rabbit ears, smoother skull forehead surfaces, precise reconstruction of missing teeth, and clear topological separation between the bird's feet and its supporting base. These enhancements are directly attributable to our robust multi-view geometric constraints, which provide reliable supervision even in challenging regions characterized by uneven illumination and sparse viewing angles. Remarkably, despite this explicit geometric supervision, our framework incurs negligible computational overhead, maintaining training efficiency strictly comparable to existing Gaussian-based approaches.

For the \textbf{Tanks and Temples (TNT)} benchmark, Table~\ref{tab:tnt_f1} summarizes the quantitative F1-score evaluations. Our method secures the highest average F1-score of \textbf{0.53}, consistently outperforming all competing methods. Specifically, it achieves the top performance on \textbf{3 out of 6} scenes and ranks second on two others, demonstrating highly robust and generalizable reconstruction quality in complex, large-scale environments.

Visual comparisons in Fig.~\ref{fig:geocompare} confirm these advantages. For instance, our method reconstructs a topologically intact \textit{Caterpillar} bucket, effectively eliminating the severe hole and depth artifacts observed in baseline methods. Furthermore, it accurately recovers the intricate pillar and associated wall details beneath the staircase in the \textit{Courthouse} scene, and captures the highly complex, hollow wheel hub structures in the \textit{Truck} scene with superior fidelity. 

Due to the higher complexity of TNT, we double the training iterations, incurring a moderate increase in runtime; results under the standard 30k iterations are also reported in Table~\ref{tab:tnt_f1}, showing a slight drop but still remaining the best overall.

\begin{figure*}[!htb]
    \centering
    \includegraphics[width=0.8\textwidth]{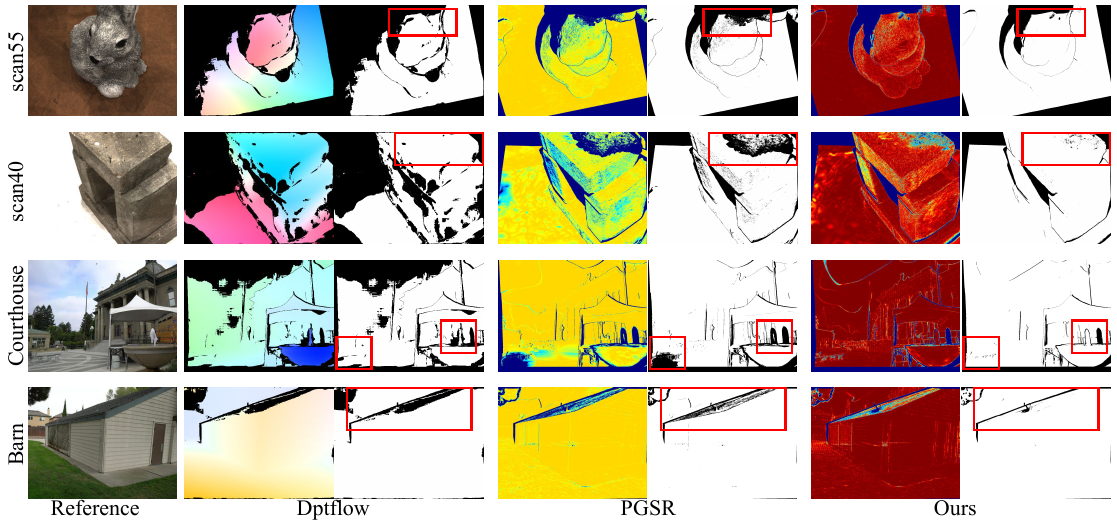}
\caption{Qualitative comparison of visibility masks across different methods.}
    \label{fig:visibility-compare}
\end{figure*}

\textbf{Visibility Comparison.}
Given the absence of ground-truth visibility masks in standard multi-view datasets, we conduct a qualitative analysis of the predicted visibility across different paradigms (Fig.~\ref{fig:visibility-compare}). We specifically compare our approach against flow-based (DPFlow~\cite{morimitsu2025dpflow}) and depth-based (PGSR~\cite{chen2024pgsr}) visibility estimation methods. Optical flow-based visibility frequently suffers from noisy and temporally unstable correspondences due to unreliable motion estimation. Conversely, depth-based visibility is inherently sensitive to initial depth inaccuracies, consistently yielding fragmented and incomplete visibility masks. In contrast, our Gaussian-level visibility modeling circumvents these bottlenecks, producing highly coherent and structurally clean masks that guarantee reliable geometric supervision strictly over valid co-visible regions.

\begin{figure}
    \centering
    \includegraphics[width=1\linewidth]{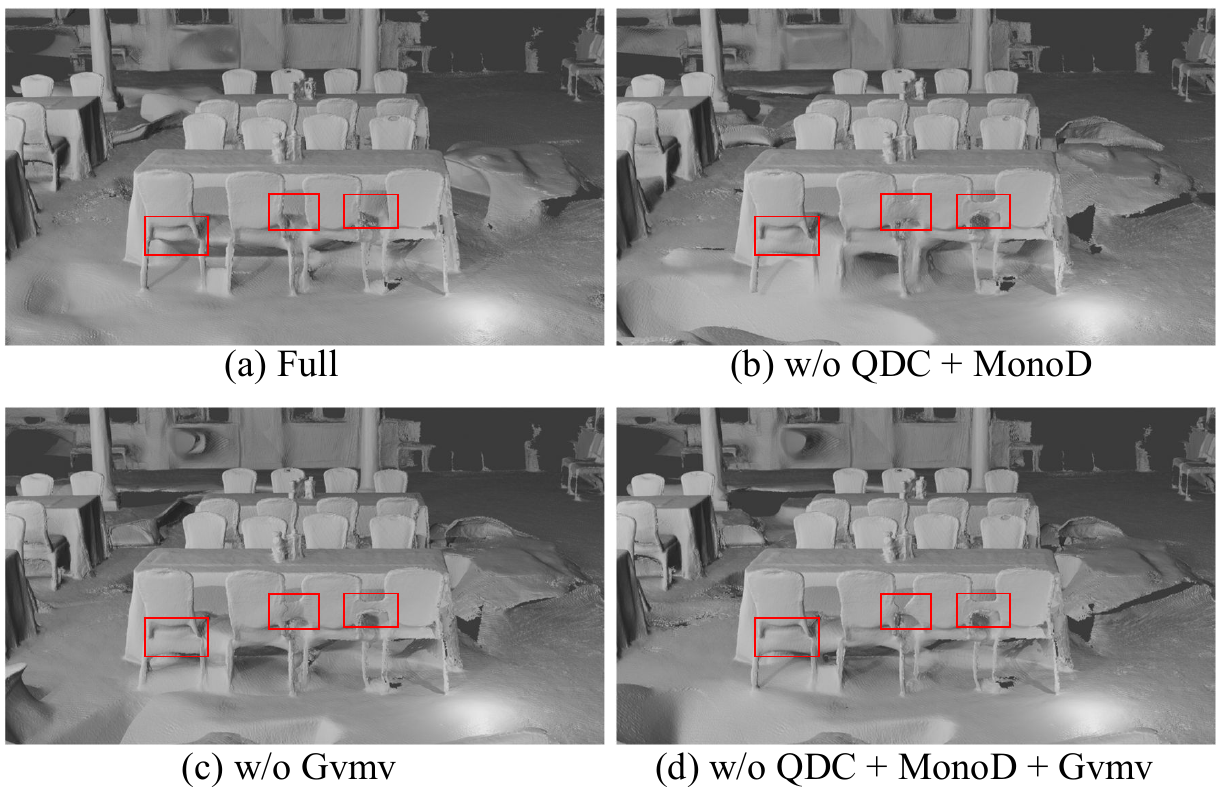}
    \caption{Qualitative ablation of the proposed components on \textit{Meetingroom}. Red boxes highlight failure cases.}
    \label{fig:meetingroom}
\end{figure}

\subsection{Ablation Studies}

We conduct comprehensive ablation studies to evaluate the individual contributions of our core components. Starting from our complete pipeline (\textbf{Full}), we systematically evaluate: 
(1) \textbf{w/o QDC}, where the progressive quadtree calibration is disabled and Gaussian depth is directly supervised by raw monocular depth~\cite{yang2024depth}; 
(2) \textbf{w/o QDC \& MonoD}, which further discards all monocular depth priors; 
(3) \textbf{w/o QDC \& MonoD \& GVMV}, which entirely removes our visibility-aware geometric consistency, effectively degrading the framework to a standard baseline; and 
(4) \textbf{w/o GVMV}, which isolates the impact of the visibility-aware formulation from the full model.

As reported in Table~\ref{tab:ablation}, omitting any component consistently degrades geometric accuracy (Chamfer Distance and F1-score) across both benchmarks. We supplement these quantitative findings with visual comparisons on the TNT \emph{Meetingroom} scene (Fig.~\ref{fig:meetingroom}). The progressive structural deterioration of the chairs under different ablation settings explicitly highlights the complementary nature of our modules, confirming that GVMV and QDC function synergistically to achieve high-fidelity surface reconstruction.
\begin{table}[!ht]
\caption{
Ablation study on DTU and TNT.
}
\label{tab:ablation}
\centering
\small
\setlength{\tabcolsep}{6pt}
\begin{tabular}{l|c|c}
\hline
Setting 
& CD (DTU) $\downarrow$
& F1 (TNT) $\uparrow$ \\
\hline
Full                                
& \textbf{0.493} & \textbf{0.530} \\

w/o QDC                            
& 0.505 & 0.520 \\

w/o QDC + MonoD                   
& 0.512 & 0.513 \\

w/o QDC + MonoD + GVMV            
& 0.519 & 0.503 \\

w/o GVMV            
& 0.511 & 0.512 \\
\hline
\end{tabular}
\end{table}

\paragraph{Sensitivity to Visibility Threshold $\tau$.} 
We analyze the robustness of our framework with respect to the visibility binarization threshold $\tau$. As shown in Table~\ref{tab:tau_ablation}, performance remains remarkably stable across a broad range of small $\tau$ values, demonstrating insensitivity to precise hyperparameter tuning. However, excessively large $\tau$ values impose overly strict filtering, prematurely discarding valid co-visible regions and degrading geometry. Furthermore, an alternative soft visibility weighting formulation yielded comparable results with no distinct empirical advantage. We thus adopt threshold-based hard gating, as binary visibility better reflects the physical notion of visibility than continuous opacity-based weighting.

\begin{table}[!t]
\centering
\caption{Ablation on the visibility threshold $\tau$. Performance is stable for small $\tau$, with the best result at $\tau=1\mathrm{e}{-4}$.}
\label{tab:tau_ablation}
\setlength{\tabcolsep}{4pt}
\resizebox{\columnwidth}{!}{
\begin{tabular}{c|cccccccc}
\toprule
$\tau$ & $1\mathrm{e}{-5}$ & $1\mathrm{e}{-4}$ & $1\mathrm{e}{-3}$ & $1\mathrm{e}{-2}$ & $1\mathrm{e}{-1}$ & $1$ & $10$ & Soft \\
\midrule
CD$\downarrow$ (DTU) & \textbf{0.493} & \textbf{0.493} & 0.497 & 0.497 & 0.497 & 0.499 & 0.503 & 0.497 \\
F1$\uparrow$ (TNT) & 0.529 & \textbf{0.530} & 0.529 & 0.529 & 0.529 & 0.526 & 0.524 & 0.529 \\
\bottomrule
\end{tabular}
}
\end{table}

\section{Conclusion}

We present a Gaussian visibility-aware multi-view geometric consistency formulation integrated with a progressive quadtree-calibrated monocular depth constraint. Together, these components establish a robust geometric supervision paradigm that significantly improves surface reconstruction fidelity.

Furthermore, our framework inherently yields high-quality multi-view visibility masks as a valuable byproduct. We anticipate these explicit cues will benefit diverse downstream applications and inspire further exploration of Gaussian-level visibility reasoning as a foundational geometric prior.

\paragraph{Limitations and Future Work.} 
Our current method lacks dedicated modeling for highly specular or transparent surfaces, where severe view-dependent effects can confound both depth prediction and visibility estimation. Extending our framework to disentangle complex material properties from view-dependent appearances remains an important avenue for future research.

  \bibliographystyle{plain}
  \bibliography{references}
%

\end{document}